\documentclass[twoside]{article}

\usepackage{microtype}
\usepackage{graphicx}
\usepackage{subfigure}
\usepackage{commath}

\usepackage{booktabs} 

\usepackage[accepted]{aistats2023}
\DeclareMathOperator*{\argmin}{arg\,min}

\usepackage{mathtools}
\usepackage{amsthm}
\usepackage[adobe-utopia]{mathdesign}

\usepackage{tabularx}
\usepackage{tikz}

\newcolumntype{M}{>{\centering\arraybackslash}m{1cm}}

\let\mathcal\undefined
\DeclareMathAlphabet{\mathcal}{OMS}{cmsy}{m}{n}

\usepackage[capitalize,noabbrev]{cleveref}

\theoremstyle{plain}
\newtheorem{theorem}{Theorem}[section]

\theoremstyle{definition}
\newtheorem{definition}[theorem]{Definition}

\theoremstyle{remark}
\newtheorem{remark}[theorem]{Remark}

\usepackage[textsize=tiny,disable]{todonotes}

\begin{document}

%

%

\twocolumn[

\aistatstitle{Counterfactual Explanations for Support Vector Machine Models}

\aistatsauthor{ Sebastian Salazar \And Samuel Denton \And  Ansaf Salleb-Aouissi }

\aistatsaddress{ Columbia University \And  Columbia University \And Columbia University } ]

\begin{abstract}
We tackle the problem of computing counterfactual explanations ---minimal changes to the features that flip an undesirable model prediction. We propose a solution to this question for linear Support Vector Machine (SVMs) models.\todo{CL} Moreover, we introduce a way to account for weighted actions that allow for more changes in certain features than others. In particular, we show how to find counterfactual explanations with the purpose of increasing model interpretability. These explanations are valid, change only actionable features, are close to the data distribution, sparse, and take into account correlations between features. We cast this as a mixed integer programming optimization problem . \\ 
Additionally, we introduce two novel scale-invariant cost functions for assessing the quality of counterfactual explanations and use them to evaluate the quality of our approach with a real medical dataset. Finally, we build a support vector machine model to predict whether law students will pass the Bar exam using protected features, and used our algorithms to uncover the inherent biases of the SVM.
\end{abstract}

\section{Introduction}
Support Vector Machines (SVMs) are  one of the best-performing discriminative models in machine learning. Despite recent advancements in areas like Deep Learning, SVMs manage to remain competitive and come with several theoretical guarantees on stability and sample-complexity. 
Support vector machines are particularly good for high dimensional data since they belong to a class of learners that learn hyperplanes with low $\ell_{2}$ norm. In general, these concept classes have low Rademacher Complexity, which means that they tend to generalize well and have a low sample complexity. This goes hand-in-hand with the ``wide margin`` property of SVMs \cite{theoryToAlgos}.\\  
To ensure responsible use of Support Vector Machines, the aforementioned theoretical guarantees are paramount --- especially in safety-critical and/or high-impact scenarios like healthcare, finance, and threat assessment. What's more, SVMs seem to perform relatively well even when data doesn't abound. This is in stark contrast to the more popular deep learning models which usually require relatively large datasets.  \\
While there is extensive theoretical research on Support Vector machines \cite{cortes1995support,vapnik1998statistical}, there is little to no information on how to use the wide-margin property to change the labels of instances with undesirable predictions and how this can be used to enhance interpretability.\todo{Added emphasis on interpretability} As an example, consider an instance with a predicted undesirable outcome (e.g., mortgage application rejected), how do we minimally change the features to flip the prediction of the original data? Explanations of this form give the decision-maker feedback on what features are most relevant to the model in the decision-making process and are known as \textbf{counterfactual explanations}. \todo{Added more emphasis on interpretability here!}
Providing explanations of this form has become increasingly important, especially in cases where automated decision-making has the potential to drastically impact human lives \cite{Wachter17}. Furthermore, legal regulations ---like the European Union's General Data Protection Regulations--- are demanding responsible deployment of machine learning models. As such, it is important to ensure that these models are used responsibly and ethically in practice.

\todo{Commented out these rhetorical questions as they are causal in nature}
In particular, we propose an integer programming formulation to finding counterfactual explanations for support vector machines. Taking advantage of the extra structure that comes with their wide-margin property, we show that counterfactual explanations from support vector machine models are stable. Additionally, evaluating the quality of counterfactual explanations is a well-known problem in the literature, and an active area of research. We propose two novel cost functions to evaluate the quality of counterfactual explanations that are scale invariant, and take on low values when the counterfactuals have desirable statistical properties. 

Previous research (e.g., \cite{eloviciB03,liu00multilevel,Ras2003IIS,trepos05}) has 
defined {\em actionability} as giving a recommendation of the \textit{easiest} way for a user to change an undesirable model prediction, based on a set of rules.\todo{Changed this...}
While rule-based classifiers allow for interpretability of the feature decisions, they are not the best performing models. 

We illustrate our approach with a \textit{diabetes} dataset, a serious disease characterized by eleveted levels of blood sugar. Over time, this can lead to serious damage to the heart, blood vessels, brain and kidneys. Diabetes has no known cure, however, in some cases (i.e., Type II diabetes), prevention can play an important role in reducing a patient's risk. This is done by managing risk factors like weight, BMI, physical activity, blood pressure, glucose levels in blood, etc. The way these factors interact is not fully understood, and controlling all of these factors simultaneously can be highly unrealistic for most patients. 

With diabetes, as in many other health problems, one challenge that a physician might face is to make important decisions about what should be done to cure a patient or reduce the risk of a specific disease. It has been increasingly common to see Machine Learning models getting incorporated into the decision-making process. As such it is paramount to be able to understand how these models make decisions. 

We define an action as a change in the value of some feature (e.g., weight of a patient, or the patient's lack of physical activity). Our approach focuses on a methodology where a datapoint is framed as a vector of features. We will then use the vector to predict an outcome based on the given SVM model. The goal will be to minimize the \textit{weighted distance} of the action that shifts the features to a point along the margin of the desired outcome. The weighted distance will be some combination of the distance traveled in each feature multiplied by a weight that accounts for the varying scales of each feature. \todo{Changed this} 

\todo{Causal language above...}

Although there is recent literature on computing counterfactual explanations for linear classifiers, we are ---to the best of our knowledge--- the first to propose an approach that takes advantage of the wide-margin property of support vector machines. 

\section{Related Work}\label{related work}

Most research on learning counterfactual explanations goes back to rule-based models,   
and can be classified into three main lines of research: 
(1) use of predefined set of simple actions that are mapped to classification rules or deviations \cite{AdomaviciusT1997,EloviciSK03,JiangWTF05,Shapiro94}; (2)  sift through a set of association or subgroup discovery rules \cite{Gamberger02,Lavrac2002ICMLWorkshop,Liu01}  to identify descriptive actionable rules
of the form $\text{Condition} \to \text{Class}$, 
(3) elaborate actions by comparing pairs of classification rules with contradicting classes 
to generate action rules \cite {Ras2002G,Ras2003IIS, Ras2005,Tsay2005}. 
Given an ``unsatisfactory prediction'', the work in \cite{trepos05,Trepos2011BuildingAF} discovers a set of action recommendations 
to improve that situation using classification rules.
Finally, actionability is formalized as a case-based reasoning problem in \cite{nearest-sv,Yang2003}. Given an example to reclassify, the authors propose to search for the closest cases with the desirable class label, among training examples,  centroids of the clusters of the opposite class, or  
SVM support vectors. Choosing the closest support vector will be used as a baseline to compare our proposed approach. 

Recent research on the interpretability and fairness of machine learning models 
also looks at how counterfactual explanations can be used to understand how a model makes decisions (e.g., \cite{Laugel2018,Laugel2017InverseCF,ustun2019,pmlr-v108-karimi20a,rawal2020i,Kenny2021OnGP,karimi2020algorithmic}). 
\todo{Deleted above} The transformed instance that results from making changes to the feature values of a given  instance in order to switch its label, is called  a nearest  {\em counterfactual} explanation or {\em actionable recourse}. 
\cite{ustun2019} frame the problem as an {\em audit} task. They formulate an optimization problem expressed as an integer program solved by an IP solver,  to uncover minimal cost actions  in linear classifiers. Other lines of work that use IP solvers to compute counterfactual explanations include \cite{Artelt21,Artelt19,Artelt21}.
\cite{Wachter17} present a comprehensive approach to counterfactual explanations without opening the black box model.  Their approach is motivated by the need for the algorithmic accountability stated in the rules of the General Data Protection Regulation (GDPR).
\todo{Commented some stuff above and also unsure what this sentence means...} In the same context of the existence of a black box model and no information about the training data, \cite{Laugel2018} identify the closest point classified differently using a growing sphere algorithms. 
The idea is to find the minimal change needed to inverse the classification. 
\todo{Don't know what to do about this since the interpretation of the weight matrix that we had is causal so I'll comment for now....}
Karimi et al. in \cite{pmlr-v108-karimi20a} propose MACE, a model-agnostic counterfactual explanations approach, that uses a sequence of satisfiability (SAT) problems to find the nearest counterfactual. MACE requires to convert the model into a characteristic First Order Logic formulas. 
However, the applicability of their approach depends on the possibility of such a conversion (which is challenging for complex non linear models) and the efficiency of a SAT solver used as an oracle.  
Algorithmic recourse proposed in \cite{karimi2020algorithmic} aims to  shift from counterfactual explanation to minimal interventions using a causal graph. In this context, the explanations of our model should be interpreted as causal when the SVM model only uses data from the parents of the target variable in the causal diagram for prediction. Additionally, it is worth noting that methods like inverse probability weighting could be used alongside our method to calculate the average causal effect. We leave the causal aspect of this line of research to future work. \todo{This is what I added on our model not being causal.}



\section{Actionability with SVMs}\label{methods}
The problem of computing robust counterfactuals for Support Vector Machines is modeled as a Mixed Integer Quadratic Program (MIQP). In this line of work, we extend the research of \cite{ustun2019}, \cite{Artelt21}, and \cite{Wachter17} in order to: (a) leverage the wide-margin property of SVMs, and (b) handle the problem of \texttt{actionability}. 

Given an instance of the training data $x \in \mathcal{X}$ with an unfavorable outcome $y \in \{\pm 1\}$, our goal is to compute the  closest counterfactual $x' \in \mathcal{X}$, with label $y' = -y$ that respects the margin constraints of the Support Vector Machine. In its most general form, this is done by solving the following optimization problem
\begin{align}
    \arg \min_{x' \in \mathcal{X}} \quad & d(x,x') \nonumber \\ 
    \textrm{ s.t. } \quad & y'\left(\langle w, x'  \rangle + b\right) \geq 0  \label{e1} \\
    & \abs{\langle w, x' \rangle + b} \geq 1. \label{margin} 
\end{align}
Where $w$ and $b$ are the coefficients of the Support Vector Machine.   \\
Now, note that the original optimization problem (\ref{e1}) is equivalent to: 
\begin{align*}
    \arg \min_{x' \in \mathcal{X}} \quad & d(x,x') \\ 
    \textrm{ s.t. } \quad & \langle w, x' \rangle + b + y \geq 0.
\end{align*}
All that remains is to choose a metric to handle the problem of actionability. We choose the weighted Euclidean distance metric to reflect the fact that features may exhibit deferent length scales. Using this metric the optimization problem becomes
\begin{align}
        \arg \min_{x' \in \mathcal{X}} \quad & \norm{W^{1/2}(x-x')}_{2}^{2} \label{OPT} \\ 
        \textrm{ s.t. } \quad & \langle w, x' \rangle + b + y \geq 0.\nonumber
\end{align}
Where $W \in \mathbb{R}^{n\times n}$ is a diagonal matrix of positive weights to be chosen by the user. From an intuitive standpoint, changes in coordinates with a higher weight are penalized more heavily and as a result the optimization problem will tend to leave those coordinates unchanged. \\ 
In practice, datasets usually have a mix of categorical and real-valued features, meaning that $\mathcal{X} \cong \mathbb{Z}^{k} \times \mathbb{R}^{n-k}$. Non-ordinal categorical features are usually one-hot encoded, and any valid counterfactual explanation should maintain this property. To handle this restriction, we could introduce constraints by considering a partition $\{J_{i}\}_{i=1}^{m}$ of the set $[k]$, where the sets $\{J_{i}\}_{i=1}^{m}$ contain the entries of a one-hot encoded feature, and enforce that $\forall i: \sum_{k \in J_{i}}x_{k} = 1$. With these additional constraints, the optimization problem becomes
\begin{align*}
        \argmin_{x' \in \mathbb{Z}^{k} \times \mathbb{R}^{n-k}} \quad & \norm{W^{1/2}(x-x')}_{2}^{2} \\ 
        \textrm{ s.t. } \quad & \langle w, x' \rangle + b + y \geq 0. \\
         \forall i\in [m]&: \sum_{s\in J_{i}}x'_{s} = 1.
\end{align*}
This is just a MIQP which we express in standard form below
\begin{align}\label{MIQP}
        \argmin_{x' \in \mathbb{Z}^{k} \times \mathbb{R}^{n-k}} \quad &\nonumber \frac{1}{2}x'^{\intercal}(2W)x' + \left(-2Wx \right)^{\intercal}x'\\ 
        \textrm{ s.t. } \quad & \langle w, x' \rangle + b + y \geq 0 \\
         \forall i\in [m]&: \sum_{s\in J_{i}}x'_{s} = 1.\nonumber
\end{align}
\section{Desirable properties of counterfactual explanations.}\label{methods:2}
The method that we have introduced so far does not have several desirable properties of counterfactual explanations. For instance, it fails to account for the relationships between input variables.\\
In general, evaluating the quality of counterfactual explanations is a well-known problem and an active area of research. In the following subsection we address how to deal with the issues of feature correlation, plausibility, sparsity, actionability, and stability ---all of which are desirable properties of counterfactual explanations. In the context of integer programming, the problem of computing counterfactual explanations with these properties has already been explored in the literature (\cite{Artelt21, Wachter17, ustun2019}. Moreover, we propose ways of adapting these ideas to support vector machines. Similarly, we introduce two novel cost functions for assessing the quality of counterfactual explanations. For a more comprehensive survey about the evaluation of counterfactual explanations, the reader is referred to \cite{Verma20}.   
\subsection{Stability}
We remark that an important property of the solutions of (\ref{e1}) is that they are stable in the sense that they are robust to perturbations. To see why this is the case, note that enforcing the wide-margin constraints of (\ref{margin}) implies that one can find an open neighborhood of the solution $x'$ ---call it $\mathcal{N}_{x'}$ --- such that for every $v \in \mathcal{N}_{x'}$ it holds that $y'\left(\langle w,v \rangle + b\right) \geq 0$ (i.e., $v$ has the same label as $x'$). This implies that, for every $v \in \mathcal{N}_{x'}$, there is a $\delta \in \mathcal{X}$ such that $v = x' + \delta$. In other words, we could perturb the counterfactual $x'$ without changing its label! More formally,
\begin{definition}
Let $\mathcal{X}$ be a Hilbert Space over $\mathbb{R}$, $\mathcal{H} \subseteq \{ f:\mathcal{X} \to \{\pm1\}\}$ be a class of functions and $h \in \mathcal{H}$ be a classifier. Let $x \in \mathcal{X}$ be such that $h(x) = y$. We say $x' \in \mathcal{X}$ is a $\delta$-stable counterfactual explanation for $x$ relative to $h$ if and only if there exists an open ball $\mathcal{B}(x',\delta)$ such that for every $v \in \mathcal{B}(x',\delta)$, we have that $h(v) = y' := -y$.  
\end{definition}
\begin{theorem}
Let $\mathcal{D} \subseteq (\mathcal{X} \times \{\pm 1\})^{m}$ be a dataset, $H \subseteq \{ x \mapsto \text{sign}(\langle x, w \rangle + b)\}$, be the concept class of linear functions, and $f \in H$ be a linear support vector machine with $\gamma = \sup_{(x,y) \in \mathcal{D}}\abs{\langle w, x \rangle + b}$. Let $x,x' \in \mathcal{X}$ be such that $f(x) = y$, $f(x') = -y 
:= y'$, and 
\begin{align}
    x' = \arg \min_{x' \in \mathcal{X}} \quad & \norm{x-x'} \nonumber \\ 
    \textrm{ s.t. } \quad & y'\left(\langle w, x'  \rangle + b\right) \geq 0  \label{qp}\\
    & \abs{\langle w, x' \rangle + b} \geq 1. \nonumber
\end{align}
\end{theorem}
then $x'$ is a $\gamma/\norm{w}$-stable counterfactual explanation for $x$. 
\begin{proof}
Note that, without loss of generality, we can always re-scale the parameters of the support vector machine to be such that 
\begin{align*}
    \gamma &= \sup_{(x,y) \in \mathcal{D}}\abs{\langle w, x \rangle + b} \\
    &= 1. 
\end{align*}
Thus, it suffices to show that the solution to \ref{qp} is a $1/\norm{w}$-stable counterfactual explanation. \\
WLOG, take $y' = 1$ and $b = 0$. Now, let $x'$ be a solution to \ref{qp}, then it follows that $\langle w, x'  \rangle  \geq 1$. Let $H = \{\xi :  \langle w, \xi \rangle \geq 0\}$ be the closed halfspace whose boundary $\partial H$ is the hyperplane defined by the vector $w$. Since $\langle x', w \rangle \geq 1$, it follows that $x' \in H\backslash \partial H$, meaning that $x'$ is an interior point of $H$. This means that there is an $\eta \in \mathbb{R}$ such that the open ball $B(x',\eta)$ is fully contained in $H$; that is $B(x,\eta) \subseteq H$. Now, take $\eta$ to be the distance of the closest point from $x'$ to the boundary of $H$ (i.e., the hyperplane $\partial H$ defined by $w$). More formally, take $\eta = d(x',\partial H) := \inf_{\partial h \in \partial H} \norm{x'-\partial h}$. Then it is clear that $B(x',\eta) \subseteq H$. However, note that $\eta$ is just the distance of the counterfactual $x'$ to the hyperplane, which is given by $1/\norm{w}$. This shows that $B(x',1/\norm{w}) \subseteq H$, meaning that every $v \in B(x',1/\norm{w})$ is such that $\langle v, w \rangle \geq 0$ which implies that $f(v) = sign(\langle v, w \rangle) = sign(\langle x', w \rangle) = 1$, completing the proof.
\end{proof}
\begin{remark}
Recall that the optimization problem of the support vector machine is given by:
\begin{align*}
    &\max \frac{1}{\norm{w}} \\
    s.t. \quad & \forall i: y_{i}\left(\langle w, x_{i} \rangle \right) \geq 1. 
\end{align*}
It is worth noting that the stability of counterfactual explanations defined by solutions to (\ref{qp}) scale like $1/\norm{w}$ ---the quantity being maximized by the SVM! Thus, it that follows that these counterfactual explanations are maximally stable. 
\end{remark}

\subsection{Accounting for correlations between features}

\cite{Artelt21} first introduced a method to account for correlations between features that and can easily be integrated to SVM-ACE. In particular, given the action of a counterfactual defined by $\delta := x' - x$, we set our new counterfactual to be $x_{cf} := x + \Sigma \delta$, where $\Sigma$ is the empirical covariance matrix. This new counterfactual $x_{cf}$ will account for linear relationships between the features. However, instead of a post-hoc application of the covariance matrix, we note that the application of the covariance matrix can be directly incorporated into the optimization problem in the following way:
\begin{align}\label{MIQP:corr}
        \argmin_{x' \in \mathbb{Z}^{k} \times \mathbb{R}^{n-k}} \quad & \nonumber \norm{W^{1/2}\Sigma^{-1/2}(x-x')}_{2} \\ 
        \textrm{ s.t. } \quad & \langle w, x' \rangle + b + y \geq 0 \\
         \forall i\in [m]&: \sum_{s\in J_{i}}x'_{s} = 1. \nonumber
\end{align}
Which is just a weighted version of the Manhalanobis distance. This distance measure has been widely studied in the literature and it has the desirable property of accounting for the correlations between features. 


\subsection{Plausibility}
When computing counterfactual explanations, it is important to ensure that they are not outliers with respect to the observed data. In other words, we want the computed counterfactual explanations to be \texttt{plausible}. In the literature, plausibility is usually enforced through the use a model that is used to learn class prototypes (i.e., a vector that is a good representative for examples with a certain label) for each of the labeled examples. In this case, use linear constraints to guarantee that the counterfactual is ``close'' to the class prototype of the class $y' = -y$. In particular, let $v_{1}$ and $v_{-1}$ be prototypes for the examples with labels of $1$ and $-1$ respectively. We enfoce the linear constraint $\abs{x' - v_{y'}} \preccurlyeq \epsilon$, which is equivalent to $\norm{x' - v_{y'}}_{\infty} \leq \epsilon$, for some hyperparameter $\epsilon \in \mathbb{R}^{+}$. With this new plausibility constraint, the new MIQP is written as:
\begin{align}\label{MIQPL:GLVQ}
        \argmin_{x' \in \mathbb{Z}^{k} \times \mathbb{R}^{n-k}} \quad & \norm{W^{1/2}(x-x')}_{2}^{2} \nonumber\\ 
        \textrm{ s.t. } \quad & \langle w, x' \rangle + b + y \geq 0. \\
         \forall i\in [m]&: \sum_{s\in J_{i}}x'_{s} = 1 \nonumber \\
        & \quad \norm{x' - v_{y'}}_{\infty} \leq \epsilon. \nonumber
\end{align}
\subsection{Sparsity}
In some cases, it is desirable that the computed counterfactuals are sparse so that a minimal number of features have to be changed to achieve a desirable outcome. To get sparse counterfactual explanations, a common approach is to choose a metric like the weighted $\ell_{1}$ norm, that is used for computing sparse solutions to optimization problems. It is well-known in the literature that the best counterfactual explanations are computed using the $\ell_{1}$ norm. In this case, the problem can be converted into an equivalent Mixed Integer Linear Program (MILP) through the introduction of auxiliary variables. In particular, consider the following modified version of (\ref{OPT}): 
\begin{align*}
        \argmin_{x' \in \mathbb{Z}^{k} \times \mathbb{R}^{n-k}} \quad & \norm{W(x-x')}_{1} \\ 
        \textrm{ s.t. } \quad & \langle w, x' \rangle + b + y \geq 0 \\
         \forall i\in [m]&: \sum_{s\in J_{i}}x'_{s} = 1. 
\end{align*}
To turn this into an equivalent MILP, it is a standard trick to introduce auxiliary variables $d_{i}$ and enforce the constraint that $\abs{\sum_{j}W_{ij}(x_{j} - x'_{j})} \leq d_{i}$. Doing this, it is straightforward to show that we obtain the following equivalent MILP: 
\begin{align*}
        \arg\min_{x',d \in \mathbb{Z}^{k} \times \mathbb{R}^{n-k}} \quad & 1^{\intercal}d \\
        \textrm{ s.t. } \quad & \langle w, x' \rangle + b + y \geq 0. \\
         \forall i\in [m]&: \sum_{s\in J_{i}}x'_{s} = 1 \\ 
         \forall i\in [n]&: \sum_{j}W_{ij}(x_{j} - x'_{j}) \leq d_{i} \\ 
         \forall i\in [n]&: \sum_{j}W_{ij}(x'_{j} - x_{j})\leq d_{i} \\ 
         \forall i\in [n]&: 0 \leq d_{i}.
\end{align*}
\subsection{Actionability}\label{actionability}
When using these methods to achieve algorithmic recourse, it is very important to account for the fact that some features may not be actionable. In this case we want to change the decision of a model by altering actionable input variables (i.e., features that are mutable or can be changed). Our method can naturally handle actionability through the (usually diagonal) weight matrix $W$. In this case, the entries $W_{ii}$ represent the penalty of suggesting changes to the $i^{th}$ feature. Since this is a minimization problem, a higher weight means that changes in that features are penalized more heavily, and as a result aren't changed as much by the optimization problem. In the extreme case where a feature is practically inactionable (e.g., someone's family history) then the weights of those features could be set to $\infty$ and will thus remain unchanged by the optimization problem. 
\subsection{Cost functions for counterfactual explanations}
We evaluate the quality of the counterfactual explanations produced by our method by using the maximum percentile shift ($f_{1}$) from \cite{ustun2019} and two novel cost functions which we call symmetric maximum log percentile shift ($f_{2}$), and the minimax percentile shift ($f_{3}$). In closed form, these cost functions are given by:  
\begin{align}
    f_{1}(x,x') &= \max_{i \in [n]}\abs{Q_{i}(x'_{i}) - Q_{i}(x_{i})}\label{f1} \\ 
    f_{2}(x,x') &= \sum_{i\in[n]} \abs{\log \left(\frac{1-Q_{i}(x'_{i})}{1-Q_{i}(x_{i})} \right)} + \abs{ \log\left(\frac{Q_{i}(x'_{i})}{Q_{i}(x_{i})}\right)}\label{f2}\\
    f_{3}(x') &= \min_{x \in S}\max_{i\in [n]}\abs{Q_{i}(x'_{i}) - Q_{i}(x_{i})}.\label{f3}
\end{align}
Where $Q_{i}$ is the percentile function of the $i^{th}$ feature.

These loss functions all exhibit a set of desirable statistical properties that make them suitable for evaluating the quality of counterfactual explanations. In particular
\begin{itemize}
    \item They all scale invariant, meaning that the values of these cost functions are unaffected by a change in of units in the features (i.e., using pounds vs. kilograms to measure weight). 
    \item $f_{1}$ measures the greatest percentage change by which the features change relative to the original instance. From a qualitative standpoint a value of $f_{1}(x,x') = 0.1$ signifies that we would need to change all of the features of a counterfactual explanation by at most $10\%$. 
    \item $f_{2}$ is a symmetrized version of the log percentile shift function introduced in \cite{ustun2019}. It increases exponentially as $Q_{j}(x_{j}) \to 1$ and $Q_{j}(x_{j}) \to 0$, which reflects the fact that changes are harder when starting off at higher or lower percentile values (i.e., at the tails of the distribution). In other words, a change from percentiles $90 \to 95$ or $10 \to 5$ is harder than a change from percentiles $50 \to 55$. Moreover, points which are pushed towards the tails of the empirical distribution will be heavily penalized (i.e., a change from percentiles $90 \to 95$ gets penalized more heavily than a change from percentiles $90 \to 80$). Therefore, high values of this cost likely correspond to outliers and thus relatively poor counterfactual explanations. For a more detailed explanation, see \cite{ustun2019}.
    \item $f_{3}$ measures the greatest percentile shift of a feature relative to some instance of the dataset. In this case, a value of $f_{3}(x') = 0.1$ would mean that we would need to change all of the features of a counterfactual by at least $10\%$ to look like some instance that is already in the dataset. Thus, high values of this cost function are an indication that the resulting counterfactual is an outlier relative to the original dataset. Moreover, for all of the observed datapoints note that $f_{3}(x) = 0$.
\end{itemize}

\section{Algorithmic recourse through counterfactuals}\label{Experiments}
\subsection{Toy example}
\begin{figure}[!htbp]
\vspace*{-0.15in}

\begin{center}
\centerline{\includegraphics[scale=0.22]{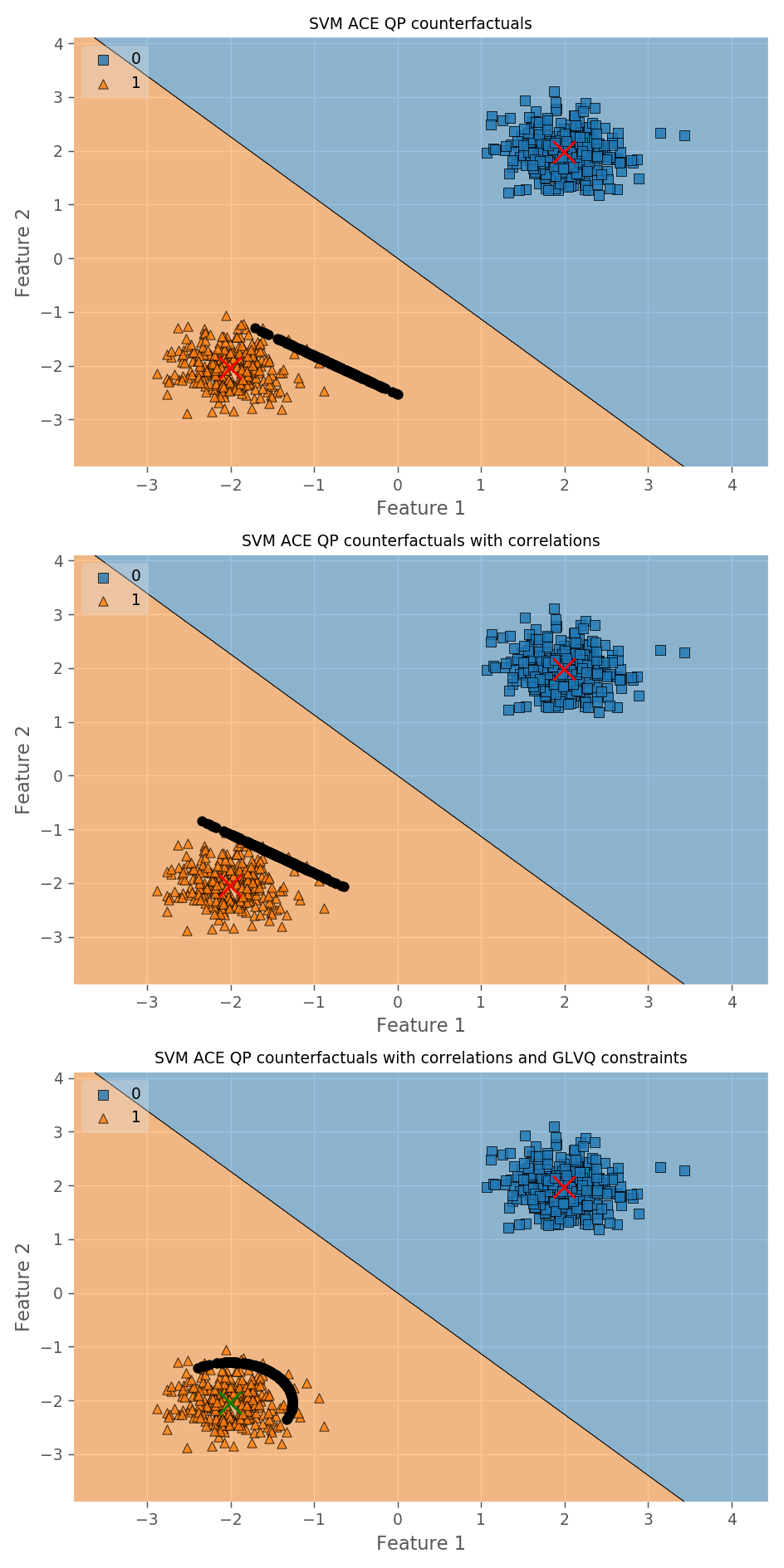}}
\caption{Example of the resulting counterfactuals (shown as black dots) of all of the examples with a label of $y=0$. (Top) Illustration of the solutions to the original optimization problem of equation \ref{MIQP}  (Middle) Illustration of the solutions to the optimization problem using the Manhalanobbis distance to account for correlations.  (Bottom) Illustration of the solutions to the optimization problem using the Manhalanobbis distance and the prototype constraints of equation \ref{MIQPL:GLVQ}. }
\label{toy data}
\end{center}
\vskip -0.2in
\end{figure}

\todo{Definitely change how the results are interpreted...} 
\todo{Add something at some point in this paper talking about individual interpretability...} 
To demonstrate our approach we simulate a separable 2-D dataset consisting of a mixture of two Gaussians. We run three simulations to qualitatively asses the effects of enforcing the plausibility and correlation constraints discussed in Sections \ref{methods} and \ref{methods:2}. The results of these experiments are shown in Figures \ref{toy data} and \ref{toy benchmarks}. 

\subsection{Diabetes dataset}
To illustrate the usefulness of our approach, we apply SVM-ACE to the Pima Indians Diabetes dataset \cite{diabetes}. It has features like the number of pregnancies the patient has had, blood pressure, BMI, insulin levels, etc. that are used to predict whether a patient is at risk of diabetes. Our goal is to suggest robust perturbabtions to the features of high-risk patients to better understand the predictions of our model. \\
We present the results of our approach in Tables \ref{diabetes cf} and asses the quality of our counterfactual explanations as compared to other methods in Section \ref{quality}. From a qualitative standpoint, our results are favorable. In particular, note that the changes for the three patients shown in Table \ref{diabetes cf} indicate that our model indicates that patients should decrease their glucose levels and BMI while slightly increasing the amount of insulin units. This provides a very clear and robust way of interpretting the predictions of our model.  Moreover, the changes suggested to flip the predictions indicate that BMI drop to the widely accepted healthy range of between 19 and 25. Similarly, a decrease in glucose concentration to  below the diabetic threshold of 140 mg/dL on most patients.

\begin{table}[!htb]
\caption{Suggested counterfactuals (CF) on actionable features for a patient to decrease the risk of Type II diabetes}
\label{diabetes cf}
\vskip 0.15in
\begin{center}
\begin{small}
\begin{sc}
\begin{tabular}{lcccr}
\toprule
Feature & Original & CF \\
\midrule
\hline
\textbf{Patient I} & & \\
Glucose (mg/dL)   &  171 & 135   \\
Diastolic BP (mm Hg) & 72 & 75  \\
Insulin   (muU/ml) & 135 & 140  \\
BMI   & 33 &  25      \\
\hline
\textbf{Patient II} && \\ 
Glucose (mg/dL)   &  155 & 110   \\
Diastolic BP (mm Hg) & 62 & 66  \\
Insulin   (muU/ml) & 495 & 501  \\
BMI   & 34 &  24       \\
\hline
\textbf{Patient III} && \\ 
Glucose (mg/dL)   &  160 & 115   \\
Diastolic BP (mm Hg) & 54 & 58  \\
Insulin   (muU/ml) & 175 & 181  \\
BMI   & 30 &  21       \\
\bottomrule
\end{tabular}
\end{sc}
\end{small}
\end{center}
\vskip -0.1in
\end{table}

\subsection{Comparison to other methods}\label{quality}
We compare the results of our approach with two methods from the literature used to compute counterfactual explanations. In particular, we compare our approach to DiCE \cite{DICE} and the nearest support vector approach from \cite{nearest-sv}.
The results of our experiments are shown in Figures \ref{toy benchmarks} and \ref{diabetes benchmarks} and Table \ref{diabetes cost}. 

\begin{table}[!htb]
\caption{Average cost of counterfactual explanations (CF) on actionable features for a diabetes patients}
\label{diabetes cost}
\vskip 0.15in
\begin{center}
\begin{small}
\begin{sc}
\begin{tabular}{lcccr}
\toprule
Method & SVM-ACE Mean Value of CF \\
\midrule
\hline
\textbf{Max percentile shift} &  \\
SVM-ACE   &  46.85   \\
SVM-ACE with corr & \textbf{46.68}  \\
Nearest SV & \textbf{38.88} \\
SVM-ACE sparse + corr & 65.46   \\
DiCE   & 78.21     \\
SVM-ACE with plausibility & 46.83 \\
\hline
\textbf{Log percentile shift} & \\ 
SVM-ACE   &  6.65   \\
SVM-ACE with corr & 6.96   \\
Nearest SV & 4.42 \\
SVM-ACE sparse + corr & \textbf{2.95}  \\
DiCE   & 12.35      \\
SVM-ACE with plausibility & 7.04 \\
\hline
\textbf{Minimax percentile shift} & \\ 
SVM-ACE   &  24.13   \\
SVM-ACE with corr & 24.26   \\
SVM-ACE sparse + corr & \textbf{23.33}   \\
DiCE   & 25.23      \\
SVM-ACE with plausibility & 24.35 \\
\bottomrule
\end{tabular}
\end{sc}
\end{small}
\end{center}
\vskip -0.1in
\end{table}

\section{Uncovering a model's bias}
Counterfactual explanations have been applied in the machine learning literature as a way of enhancing interpretability (see \cite{Verma20}). In this section, we use our methods from Sections \ref{methods} and \ref{methods:2}  to uncover the inherent biases of support vector machine models. As a case study, we use the reduced version of the LSAT dataset, to predict whether law students will pass the bar, based on undergraduate GPA, LSAT scores, and race as a sensitive attribute. We compute counterfactual explanations as solutions to equation (\ref{MIQP:corr}), for all of the students that were predicted to fail the bar exam according to our SVM model. We report the mean changes of these counterfactual explanations for 884 students in Table \ref{lsat-table}. From a qualitative standpoint, we observe that the counterfactual explanations suggest higher GPAs and LSAT scores to flip the prediction of outcomes. At the same time we see that, on average, there were around 6.2\% more candidates whose prediction would have been changed had the ``Race (white)`` attribute been set to true. These results suggest an inherent bias towards a protected feature (i.e., race) for our SVM ---in agreement with previous interpretability experiments from the literature on this dataset \cite{Wachter17}. \\ 
We compare our approach with SHAP (SHapley Additive exPlanations, \cite{SHAP17}) ---a standard, widely used, approach for interpretability of machine learning models. The shap values depict the feature importance of the SVM model and are shown in Table \ref{shap-table}. When comparing the relative importance of race as a feature in making negative predictions, we see that the features Race (Black) and Race (Hispanic) were the most negatively affected by the predictions. These results are in qualitative agreement with the results of our approach shown in Table \ref{lsat-table}. \\ 
\begin{table}[!htb]
\caption{Mean of the suggested change in features for 884 law students predicted to fail the bar exam. The support vector machine suggested that the law students change the value of the race attribute to white 6.2\% more examples relative to the other race features. The changes for categorical values are expressed as relative percentages. For continuous features, the changes are expressed in absolute terms.}
\label{lsat-table}
\vskip 0.15in
\begin{center}
\begin{small}
\begin{sc}
\begin{tabular}{lcccr}
\toprule
\textbf{Counterfactual explanations} & \\ 
\midrule
Feature & SVM-ACE Mean Change  \\
\hline
GPA    &   0.57   \\
LSAT Score & 24.9   \\
Race (White) & 6.2 \%   \\
Race (Black)   & (4.8)\%      \\
Race (Hispanic) & (1.6)\%  \\ 
Race (Other) & (0.0)\% \\ 
Race (Asian) & (0.2)\% \\
\hline 
\bottomrule
\end{tabular}
\end{sc}
\end{small}
\end{center}
\vskip -0.1in
\end{table}

\begin{table}[!htb]
\caption{Mean SHAP values for 884 students predicted to fail the bar exam.}
\label{shap-table}
\vskip 0.15in
\begin{center}
\begin{small}
\begin{sc}
\begin{tabular}{lcccr}
\toprule
\textbf{SHAP Values} & \\
\midrule
Feature & SHAP value  \\

\hline
GPA    &   0.13 \\
LSAT Score & 0.28   \\
Race (White) & (0.04)   \\
Race (Black)   & (0.14)     \\
Race (Hispanic) & (0.08) \\ 
Race (Other) & (0.01) \\ 
Race (Asian) & (0.03) \\
\hline 
\bottomrule
\end{tabular}
\end{sc}
\end{small}
\end{center}
\vskip -0.1in
\end{table}
\section{Conclusion and Discussion}\label{conclusion}
The task of computing actionable counterfactual explanations for discriminative models has been relatively unexplored in modern machine learning research. We introduce a simple, yet effective approach for using Support Vector Machines to compute  counterfactual explanations using mixed integer programming. These methods compute counterfactuals that are valid, actionable, plausible, stable, and take into account correlations between features.  \\
We also introduce two novel cost functions to evaluate the quality of counterfactual explanations. These cost functions are scale invariant and attain low values for counterfactual explanations that have favorable statistical properties relative to the data distribution. Moreover, we find that our approach outperforms other standard approaches for computing counterfactual explanations from the literature. \\
Using the counterfactual explanations, we explored the interpretability of SVM models on the LSAT dataset. In particular, we used our approach to uncover the inherent biases of the SVM model. We find that our results are both consistent with the interpretability literature, and qualitatively agree with the results produced by other interpretability methods from the literature. \\ 
To illustrate the usefulness of this approach we conducted experiments on a diabetes dataset. Our goal was to suggest actionable changes to the features of high-risk patients to decrease their risk of diabetes. When applying our approach in this context, we must note that the counterfactual recommendations are only as good as the SVM model itself. Even then, this model is limited by the fact that it fails to account for causal relationships between features. As such, this approach should always be paired with domain knowledge from experts and one should be wary about drawing causal conclusions using this method. We view the adaptation of this method for causal inference as an interesting direction for future work.  
\newpage

\newpage


\bibliographystyle{plain} 
\bibliography{refs} 

\newpage

\end{document}


%

%

\onecolumn
\aistatstitle{Supplementary material}
\section{Appendix: Value of counterfactuals on individual examples.}
\begin{figure}[!htb]
\begin{center}
\centerline{\includegraphics[width=7cm, height=15cm]{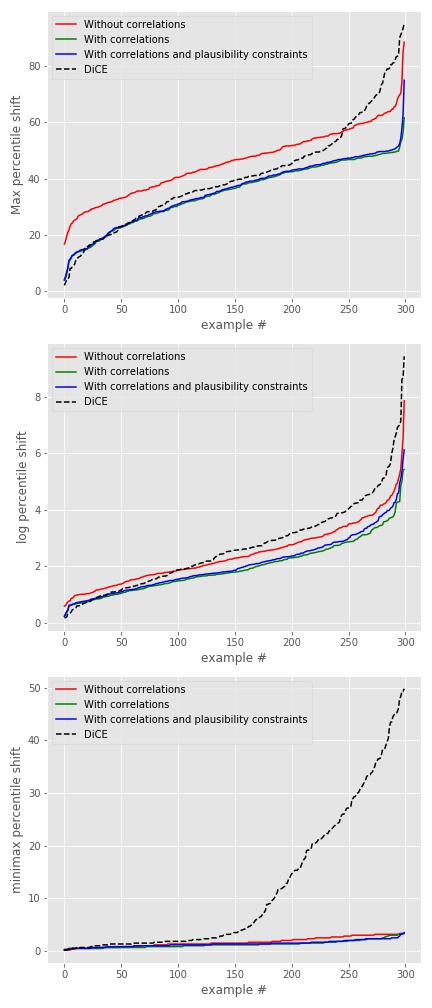}}
\caption{Example of quality of counterfactuals on the toy dataset according to the cost functions of equations 8, 9, and 10.}
\label{toy benchmarks}
\end{center}
\end{figure}

\newpage
\begin{figure}[!htb]
\begin{center}
\centerline{\includegraphics[scale=0.60]{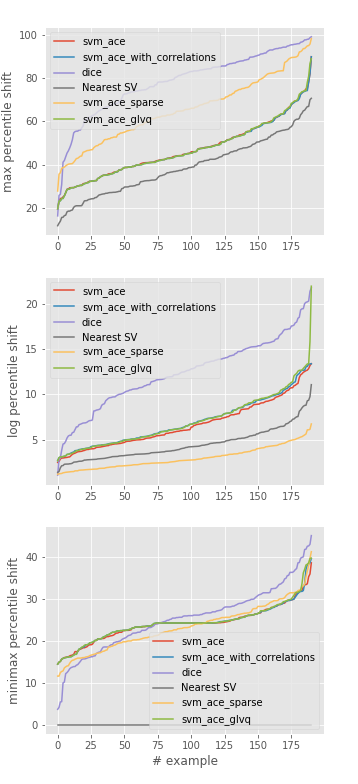}}
\vskip -0.1in
\caption{Example of quality of counterfactuals on the diabetes dataset according to the cost functions of equations 8, 9, and 10.}
\label{diabetes benchmarks}
\end{center}
\vskip -1.0in
\end{figure}
\vfill